\documentclass[runningheads]{llncs}
\usepackage{graphicx}
\usepackage{hyperref, xcolor}
\usepackage{bbm}
\usepackage{amssymb}
\usepackage{amsmath}
\usepackage{mathtools}
\usepackage{subfig}
\usepackage{tablefootnote}
\usepackage{threeparttable, booktabs}
\usepackage{url}
\usepackage[normalem]{ulem}

\usepackage{amsmath}
\DeclareMathOperator*{\argmax}{arg\,max}
\newcommand{\card}[1]{\left \vert #1 \right \vert}
\newcommand\dsum{\displaystyle\sum\limits}

\newcommand{\update}[1]{\textcolor{black}{#1}}

\begin{document}
\title{Test-time image-to-image translation ensembling improves out-of-distribution generalization in histopathology}
\titlerunning{Test-time image-to-image translation ensembling}
%


\author{Marin Scalbert\inst{1, 2} \and
Maria Vakalopoulou\inst{1} \and
Florent Couzinié-Devy\inst{2}}
\authorrunning{M. Scalbert et al.}
\institute{MICS, CentraleSupélec - Université Paris-Saclay, Gif-sur-Yvette, France \\
\email{\{marin.scalbert, maria.vakalopoulou\}@centralesupelec.fr}
\and
VitaDX International, Paris, France \\
\email{f.couzinie-devy@vitadx.com}}

%
\maketitle              
\begin{abstract}
Histopathology whole slide images (WSIs) can reveal significant inter-hospital variability such as illumination, color or optical artifacts. These variations, caused by the use of different protocols across medical centers (staining, scanner), can strongly harm algorithms generalization on unseen protocols. This motivates the development of new methods to limit such \update{loss of generalization}. In this paper, to enhance robustness on unseen target protocols, we propose a new test-time data augmentation based on multi domain image-to-image translation. It allows to project images from unseen protocol into each source domain before classifying them and ensembling the predictions. This test-time augmentation method results in a significant boost of performances for domain generalization. To demonstrate its effectiveness, our method has been evaluated on two different histopathology tasks where it outperforms conventional domain generalization, standard/H\&E specific color augmentation/\update{normalization} and standard test-time augmentation techniques. \update{Our code is publicly available at ~\url{https://gitlab.com/vitadx/articles/test-time-i2i-translation-ensembling}}.

\end{abstract}
\section{Introduction}

Histopathology is the gold standard to diagnose most types of cancer. Sampled tissues are processed following specific protocols (fixation, staining, scanning) to obtain whole slide images (WSIs). It enables tissue types/cells structures differentiation and highlights abnormalities or cancer indicators. However, these protocols substantially vary across different medical institutions resulting in significant inter-hospital variability at the WSIs level. Such variability originating from illumination and/or color variations or optical artifacts can even occur within the same hospital~\cite{lafarge2017domain} as well as over time~\cite{stacke2020measuring}. 

Intra and inter-hospital variability prevent models from generalizing well on unseen hospitals~\cite{tellez2019quantifying}. This makes domain generalization one of the primordial and studied task of computational pathology. To circumvent the drop of generalization performances in histopathology applications, recent methods have relied on standard color augmentations~\cite{tellez2019quantifying}, H\&E specific data augmentations ~\cite{tellez2018whole,faryna21,wagner2021structure}, stain normalization~\cite{reinhard2001color,macenko2009method,tellez2019quantifying,shaban2019staingan} or domain generalization \update{(DG)} techniques~\cite{sun2016deep,arjovsky2019invariant,shi2021gradient,sagawa2019distributionally}. In histopathology, data augmentation has proven to be one of the most efficient and simple technique to close the gap due to domain shift~\cite{tellez2019quantifying}. Some data augmentation/unsupervised domain adaptation (UDA) methods are even exploiting image-to-image translation methods to learn stain-invariant models~\cite{wagner2021structure,vasiljevic2021towards}. 
\update{However, UDA methods need access to unlabeled data from the target domain which is not the case in the DG setting and must be trained every time we want to predict on a new target domain. }
Currently, data augmentation methods only exploit image-to-image translation models at training which as shown by our study is not optimal.

In this paper, we propose for the first time, the use of \update{image-to-image translation} for test-time augmentation (TTA) designing a tailored ensemble strategy. In particular, the method is based upon the multi-domain image-to-image translation model StarGanV2~\cite{choi2020StarGanV2}. At test-time, it projects images from unseen domains to the source domains, classify the projected images and ensemble their predictions. 
Additionally, several ensembling strategies have been explored. The proposed TTA does not rely on any prior on the domain shifts, learning them directly from the training data. Our method, operating at test-time, can be easily combined with other \update{DG} and/or data augmentation techniques. To demonstrate its effectiveness, it has been evaluated on two different histopathology tasks and has shown better generalization over previous conventional DG, color augmentation/\update{normalization}, and standard TTA techniques \update{\cite{sun2016deep,arjovsky2019invariant,sagawa2019distributionally,shi2021gradient,cubuk2020randaugment,tellez2018whole}}.


\section{Method}


In the problem of multi-source DG, we are given $S$ source domains $\{\mathcal{D}_1, \hdots, \mathcal{D}_S\}$. Each source domain $\mathcal{D}_i$ with domain label $d_i$ contains $n_{\mathcal{D}_i}$ labeled examples $\{(x_j, {y_j}) \mid 1 \leq j \leq n_{\mathcal{D}_i}\}$. The goal is to learn a robust model from the $S$ labeled source domains so that it generalizes well on an unseen target domain $\mathcal{D}_T$. In the following sections the different components of our method are detailed.

\subsection{StarGanV2 Architecture} 

\begin{figure}
\centering
  \includegraphics[width=1\linewidth]{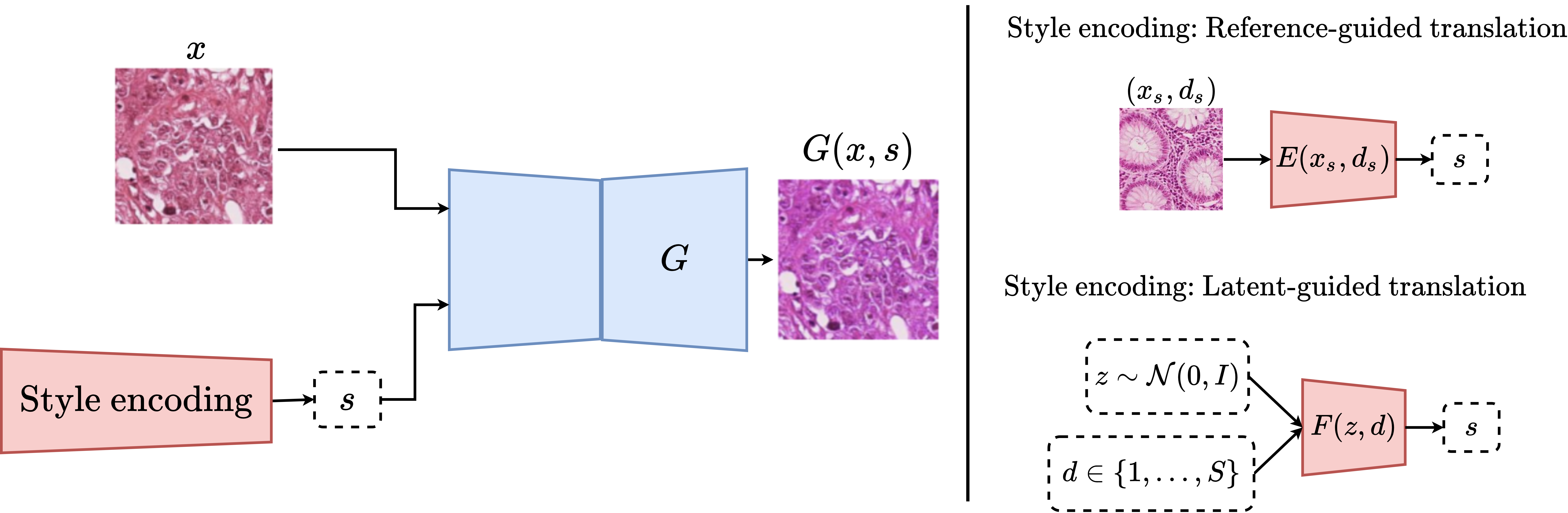}
  \captionof{figure}{The two different style encodings: on right top the reference-guided and right bottom the latent-guided image-to-image translation are presented.}
  \label{fig:ref_lat_guided_synthesis}
\end{figure}

StarGanV2 is a multi-domain image-to-image translation model capable of translating an input image $x$ into any source domain without requiring the input image domain label: this is what allows us to use it on \update{unseen} target domain. It is considered as a style based image-to-image translation model meaning that the generator takes as inputs the image that needs to be translated but also a domain-dependant style vector \update{helping the generator to translate} the input image into the corresponding domain. Style vectors can be extracted from a reference style image or from a random latent code (~\autoref{fig:ref_lat_guided_synthesis}).

More specifically, given an image $x$ and a style vector $s$ from a source domain $d$, the generator $G$ translates the image $x$ into an image $G(x, s)$ reflecting visual attributes of the domains $d$. $s$ can be computed either from a style encoder $E$ given a reference style image with its domain $(x_s, d_s)$ (i.e $s = E(x_s, d_s)$) or from a mapping network $F$ given a random latent code $z$ along with a domain label $d$ (i.e $s = F(z, d)$). Our proposed TTA method relies on latent-guided image-to-image translation due to the lighter size and lower computational overhead of $F$ compared to $E$. Additionally, some of the explored ensembling strategies exploit the different domain discriminators used during the StarGanV2 training. To better preserve the structure of the input image in the translated images, the perceptual domain invariant loss~\cite{huang2018multimodal} is added to the overall loss of StarGanV2. Examples of translated images are provided in \update{Fig. 1 of supplementary material}. 

\subsection{Test-time image-to-image translation ensembling}
\label{subsec:tta_image_to_image_translation_ensembling}

\begin{figure}
    \centering
    \includegraphics[width=\textwidth]{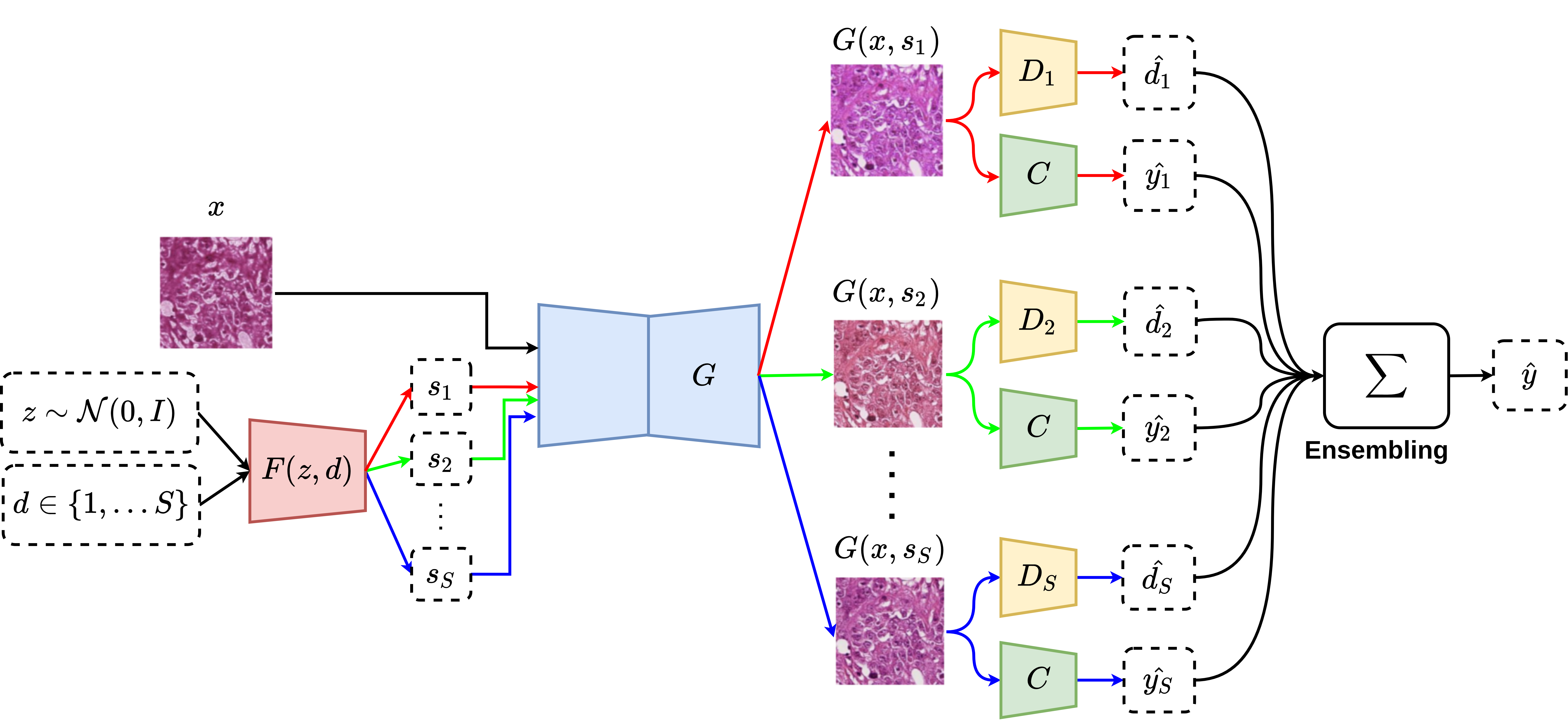}
    \caption{Test-time data augmentation with latent-guided StarGanV2 image-to-image translation}
    \label{fig:stargan_tta_diagram}
\end{figure}

\update{Once a StarGanV2 and a classifier $C$ have been trained separately on the source data, our TTA can be used at inference.} An overview of our TTA method is presented in ~\autoref{fig:stargan_tta_diagram}. Given a test image $x$, we sample $S$ random latent code vectors $\{z_1, \hdots, z_S\} \sim \mathcal{N}(0, I)$ with respective domain label $\{d_1, \hdots, d_S\}$. Each latent code vector along with its domain label are fed to the mapping network $F$ to produce the different style vectors $\{s_1, \hdots, s_S\} =\left \{F(z_1, d_1), \hdots, F(z_S, d_S) \right\}$. Then, the image $x$, along with each of the style vectors $\{s_1, \hdots, s_S\}$, \update{are} fed to the StarGanV2 generator $G$ translating $x$ into the images $\{x_1, \hdots, x_S\} = \left \{G(x, s_1), \hdots, G(x, s_S) \right\}$. From each $x_i$ that should have now similar characteristics as real images from domain $d_i$, we compute the predictions of the classifier $\hat{y}_i = C(x_i)$ and the different StarGanV2 domain discriminators score $\hat{d}_i = D_i(x_i)$ reflecting how well the image $x$ has been projected into domain $d_i$. Finally, from the classifier predictions $\{\hat{y}_1, \hdots, \hat{y}_S\}$ and the domain discriminators score $\{\hat{d}_1, \hdots, \hat{d}_S\}$, we propose three different strategies to ensemble the predictions into a final prediction $\hat{y}$. 

\textbf{Naive ensembling.}
In this method, domain discriminators scores $\{\hat{d}_1, ..., \hat{d}_S\}$ are discarded and the predictions $\{\hat{y}_1, \hdots, \hat{y}_S\}$ are simply averaged. 

\textbf{Discriminator based top-$k$ ensembling.}
A domain discriminator score should reflect how well a test image has been projected into the corresponding source domain. Following this intuition, we propose to ensemble only $k$ predictions associated to the images with the top-$k$ domain discriminators scores:
\begin{equation}
    \hat{y} = \displaystyle \frac{1}{S'}\sum_{s=1}^{S'} \hat{y}_s \quad ; \quad S' = \argmax_{\substack{I \subseteq \{1, \hdots, S\} \\ \card{I} = k}} \dsum_{i \in I}^{} \hat{d}_i
\end{equation}

\textbf{Discriminator based weighted ensembling.} In the discriminator based top-$k$ ensembling, although the predictions associated to the lowest domain discriminator scores have been discarded from the ensembling, the $k$ kept predictions contribute evenly even if there might be significant score disparities. 
Preferably, we would like to consider all the predictions to unlock the full power of \update{ensembling} while reducing the importance of predictions associated to badly projected images. That is why, we propose here to ensemble all the predictions $\{\hat{y}_1, \hdots, \hat{y}_S\}$ with respective weights $\{\alpha_1, \hdots, \alpha_S\}$ based on domain discriminators scores $\{\hat{d}_1, \hdots, \hat{d}_S\}$:
\begin{equation}
    \hat{y} = \dsum_{s=1}^{S} \alpha_s \hat{y}_s \quad ; \quad \alpha_{s} =  \dfrac{e^{\displaystyle \tfrac{\hat{d}_s}{T}}}{ \dsum_{j = 1}^{S} e^{ \tfrac{\hat{d}_j}{T}}} \quad \forall s \in \{1, \hdots, S\}
    \label{eq:disc_based_weighted_ensembling}
\end{equation}
The temperature $T$ in Eq.~\ref{eq:disc_based_weighted_ensembling} is a hyperparameter controlling how peaky the predictions weights distribution should be. When $T \rightarrow \infty$, predictions weights tend to be uniform $(\alpha_s = \dfrac{1}{S})$ which is equivalent to the naive ensembling. While, when $T \rightarrow 0$, predictions weights tend to be a one-hot encoded vector which is equivalent to the discriminator based top-$k$ ensembling with $k=1$.

\section{Experiments}
\subsection{Datasets}
\label{subsec:datasets}

\subsubsection{Patch classification of lymph node section WSIs.} For this DG task, we use the standard Camelyon17 WILDS dataset \cite{koh2021wilds} containing patches of WSIs of lymph node sections from patients with potentially metastatic breast cancer. WSIs have been collected from 5 medical centers and splits have been made with respect to the medical center. The goal of the task is to predict whether or not patches contain tumorous tissue and to generalize well on patches from unseen centers. The distribution (\#hospitals, \#WSIs, \#patches) for the \texttt{train}, \texttt{id\_val}, \texttt{val} and \texttt{test} splits are respectively $(3, 30, 302436)$, $(3, 30, 33560)$, $(1, 10, 34904)$ and $(1, 10, 85054)$. \texttt{train} and \texttt{id\_val} hospitals are identical while \texttt{val} and \texttt{test} hospitals are both distinct and different from \texttt{train} and \texttt{id\_val} hospitals. To be able to exploit the full potential of ensembling, we decide to choose the different WSIs as the different source domains $(S=30)$ rather than the number of distinct hospitals. Moreover, considering WSIs as domain can be convenient when you have no more information than the WSI itself.

\subsubsection{Tissue type classification in colorectal histological images. }
For this DG task, we use $3$ different datasets of colorectal cancer histological images namely: Kather16~\cite{kather2016multi}, Kather19~\cite{kather2019predicting} and CRC-TP~\cite{javed2020cellular}. Kather16 contains $5000$ patches from 10 H\&E
WSIs spread into 8 classes. Kather19 includes $100000$ patches from several H\&E WSIs spread into 9 classes. CRC-TP comprises a total of 196000 patches extracted from 20 H\&E WSIs and spread over 7 classes. Given that the hospital or slide information are not provided for Kather19 and CRC-TP, we only use Kather16 as the train set and Kather19 with CRC-TP as test sets. Because class definitions can vary from one dataset to another, we follow the class grouping suggested by~\cite{abbet2021self}. Additionally, we remove the class \textit{complex stroma} from Kather16 and CRC-TP which do not share the same definition. Finally, our problem consists in classifying patches from colorectal cancer histological images into 7 classes (\textit{tumor}, \textit{normal}, \textit{stroma}, \textit{lymphocytes}, \textit{debris}, \textit{adipose} and \textit{background}) and to generalize well on unseen domains (Kather19, CRC-TP). Since our training set (Kather16) does not provide hospital information for each patch, we considered WSIs as source domains $(S=10)$.

\subsection{Method evaluation}
\label{subsec:results}
\begin{table}[t]
    \centering
    \resizebox{\textwidth}{!}{
    \begin{tabular}{|l|c|c|c|c|c|}
    \hline
    & \multicolumn{3}{c}{Lymph node patch classification} & \multicolumn{2}{|c|}{Colorectal tissue type classification} \\
    \hline
                                                    &  \texttt{id\_val}  &  \texttt{val}  &  \texttt{test}  &  Kather19  &  CRC-TP  \\
    \hline
                     Base                 &   $\mathbf{98.5 (0.1)}$    &  $80.3 (4.4)$  &  $62.7 (7.4)$   &    $70.8 (1.8)$     &   $43.6 (3.7)$   \\
    \hline
    \hline
    \textbf{SOTA domain generalization methods} &  &  &  &  & \\
                         CORAL~\cite{sun2016deep}                      &                   &  $86.2 (1.4)$  &  $59.5 (7.7)$   &                     &                  \\
                          IRM~\cite{arjovsky2019invariant}                       &                   &  $86.2 (1.4)$  &  $64.2 (8.1)$   &                     &                  \\
                       Group DRO~\cite{sagawa2019distributionally}                    &                   &  $85.5 (2.2)$  &  $68.4 (7.3)$   &                     &                  \\
                          FISH~\cite{shi2021gradient}                      &                   &  $83.9 (1.2)$  &  $74.7 (7.1)$   &                     &                  \\
    \hline
    \hline
    
    \textbf{\update{Color normalization}} &  &  &  &  & \\
        \update{Macenko~\cite{macenko2009method}} & \update{} & \update{81.6(1.2)} & \update{92.5(1.8)} & \update{64.9(3.3)} & \update{51.6(1.2)}\\
    \hline
    \hline
    \textbf{Train-time data augmentation} &  &  &  &  & \\
              Base + RandAugment~\cite{cubuk2020randaugment}          &                   &  $90.6 (1.2)$  &  $82.0 (7.4)$   &                     $72.6 (3.2)$     &   $51.0 (5.1)$                  \\
      Base + H\&E staining color jitter~\cite{tellez2018whole} &                   &  $88.0 (4.2)$  &  $91.6 (1.9)$   &                     $72.4 (3.3)$     &   $47.6 (4.8)$                  \\
                Base + StarGanV2 data aug            &   $\mathbf{98.4 (0.0)}$    &  $89.6 (0.7)$  &  $76.4 (4.5)$   &    $66.3 (3.3)$     &   $37.7 (6.1)$   \\
    \hline
    \hline
    \textbf{Test-time data augmentation} &  &  &  &  & \\
        Base + StarGanV2 data aug + geometric TTA    &   $\mathbf{98.5 (0.1)}$    &  $90.0 (0.6)$  &  $76.5 (4.2)$   &    $66.2 (3.5)$     &   $37.6 (5.9)$   \\
      Base + color jittering + color jittering TTA   &   $97.4 (0.1)$    &  $91.2 (0.4)$  &  $77.2 (1.2)$   &    $\mathbf{74.5 (2.5)}$     &   $\mathbf{51.7 (6.1)}$   \\
      \hline
    \hline
    \textbf{Our method} &  &  &  &  & \\
          Base + StarGanV2 data aug + naive ens      &   $96.9 (0.3)$    &  $\mathbf{92.8 (0.2)}$  &  $\mathbf{94.0 (1.2)}$   &    $\mathbf{72.9 (2.0)}$     &   $\mathbf{54.4 (3.2)}$   \\
     Base + StarGanV2 data aug + disc top-k ens   &   $97.3 (0.2)$    &  $92.1 (0.2)$  &  $\mathbf{94.0 (1.2)}$   &    $\mathbf{72.9 (2.0)}$     &   $\mathbf{54.4 (3.2)}$   \\
     Base + StarGanV2 data aug + disc weighted ens &   $97.4 (0.2)$    &  $\mathbf{92.7 (0.2)}$  &  $\mathbf{94.0 (1.2)}$   &    $\mathbf{72.9 (2.0)}$     &   $\mathbf{51.6 (3.3)}$   \\
    \hline
    \end{tabular}
    }
    \caption{\update{Accuracy on Camelyon17 WILDS, weighted F1-score on Kather19 and CRC-TP.} \textit{Naive ens}, \textit{disc top-$k$ ens} and \textit{disc weighted ens} referred respectively to naive ensembling, discriminator based top-$k$ ensembling and discriminator based weighted ensembling. Best performances are highlighted in \textbf{bold} and standard deviations are specified in parenthesis.}
    \label{tab:all_perfs}
\end{table}

In this section, the proposed TTA \update{with the different} ensembling methods are evaluated on the two different tasks. We report a baseline (base) corresponding to cross-entropy minimization on the source domains. Additionally, we also evaluate a method named StarGanV2 data aug where StarGanV2 is used only for train-time data augmentation. The performances on Camelyon17, Kather19, CRC-TP are respectively averaged over $10$, $5$ and $5$ independent runs and reported on \autoref{tab:all_perfs}. Implementation details about the StarGanV2 and classifier are provided in \update{Table 1 of supplementary material}.


\subsubsection{State-of-the-art results on Camelyon17 WILDS}
Our method obtains state-of-the-art results on Camelyon17 WILDS. \update{Results of competing methods are reported in the "SOTA domain generalization methods", "Color normalization" and the two first lines of the "Train-time data augmentation" row.} On the \texttt{val} and \texttt{test} splits, our TTA method with naive ensembling outperforms the previous best method by $+2.2\%$ and 
\update{$+1.5\%$} respectively. It is worth mentioning that our method learns the domains variations from the data while the two previous best methods on \texttt{test} use expert priors  (H\&E staining color jittering and \update{Macenko color normalization}). Therefore, in theory, our method could generalize on other medical imaging modalities. Compared to the best DG method FISH, that do not use any priors on the domain change, our method leads to much better generalization with $+8.9\%$ and $+19.3\%$ accuracy improvements. The TTA method performs also well on source domain data (\texttt{id\_val} split) even if performances are slightly worse than the baseline.

\subsubsection{Better robustness on Kather19 and CRC-TP}
The second task is more challenging as only~4400 images are available in Kather16 to train the StarGanV2 and the classifier. The experiments on these datasets have two objectives: show that our method generalizes on another problem/dataset and that it can work even with relatively small size datasets. 
Previous works usually use Kather16 for evaluation making \update{performances} comparisons impossible. However, the results on Kather19 and CRC-TP shows that our method increases the baseline results significantly: $+2.1\%$ and $+11.8\%$ respectively. These improvements are not as large as in Camelyon17 which is probably due to a less robust StarGanV2 model. This hypothesis is reinforced by the bad performances of train-time data augmentation using the StarGanV2. 

\subsection{Ablation studies}

\subsubsection{Analyzing data augmentation effects}
We have studied independently the effects of the train-time and test-time data augmentations based on the same StarGanV2.
The train-time data augmentation with StarGanV2 alone helps to make the model more robust on Camelyon17 ($+9.3\%$ and $+13.7\%$ for \texttt{val} and \texttt{test} splits). However, on Kather16, where the dataset is small, it leads to loss of performances ($-4.5\%$ and $-5.9\%$). On both tasks, when train-time data augmentation based on StarGanV2 is combined with our TTA method, generalization performances improve considerably.

To evaluate the impact of our TTA, geometric TTA and color jittering TTA have been evaluated.
Both perform quite well: in particular, the color jittering prior seems to be correct for $3$ out of the $4$ target datasets making it as good as our method on the colorectal datasets and nearly as good on the Camelyon17 \texttt{val} split. However, color jittering TTA fails on the Camelyon17 \texttt{test} split ($-16.8\%$ compared to our TTA method). This suggests that the augmented target domain distribution is not overlapping enough the source domains distributions and that projecting images onto source domains is a better approach to achieve generalization. \update{A t-SNE~\cite{van2008visualizing} plot provided in \update{Fig. 2 of} supplementary material confirms that projecting target examples onto the different source domains is one of the key to achieve better generalization.}

\subsubsection{Comparison of the ensembling strategies}
\begin{figure}[t]
    \centering
    \includegraphics[width=0.33\textwidth]{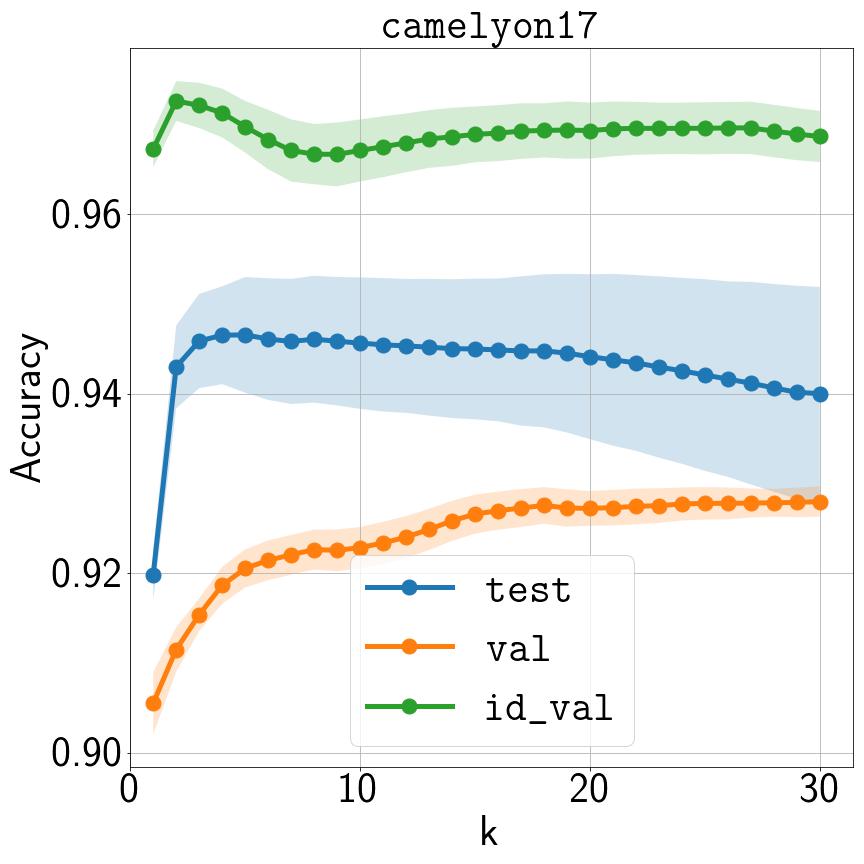}
    \includegraphics[width=0.66\textwidth]{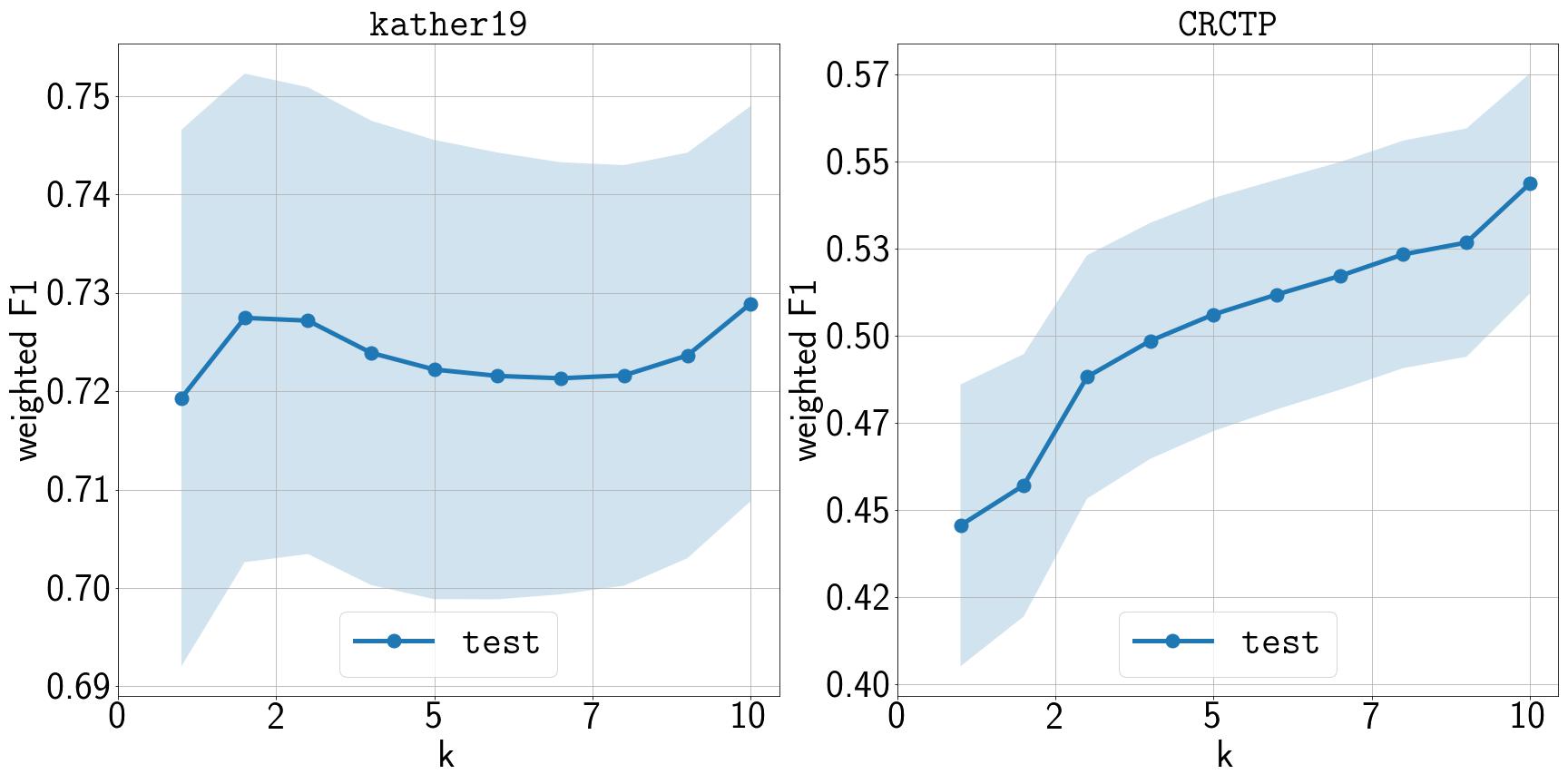}
    \includegraphics[width=0.33\textwidth]{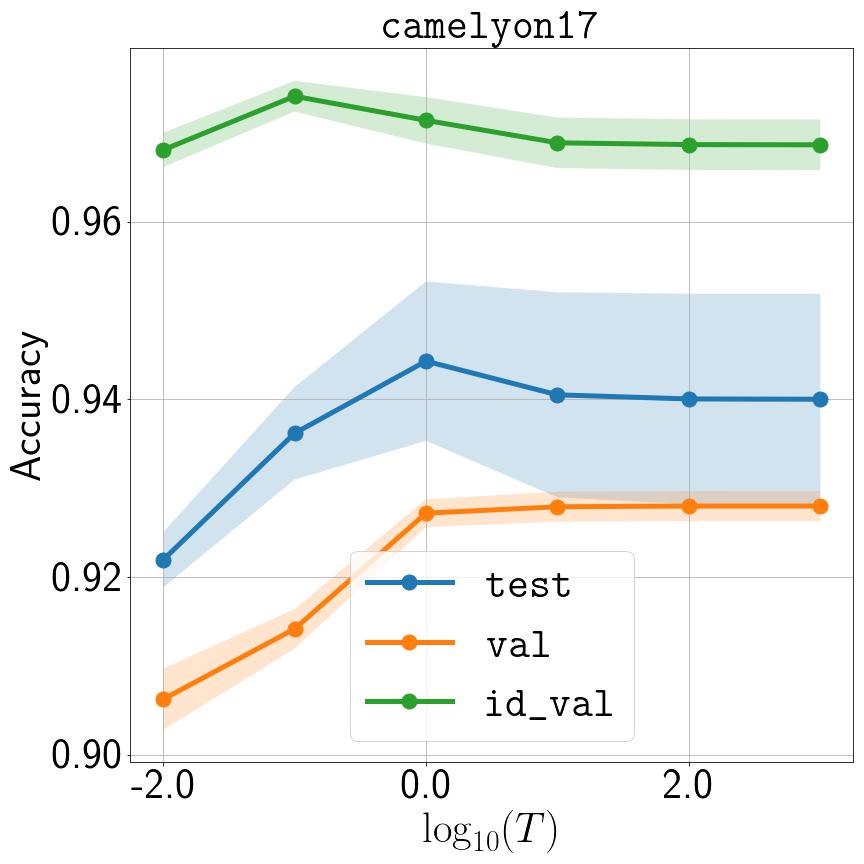}
    \includegraphics[width=0.66\textwidth]{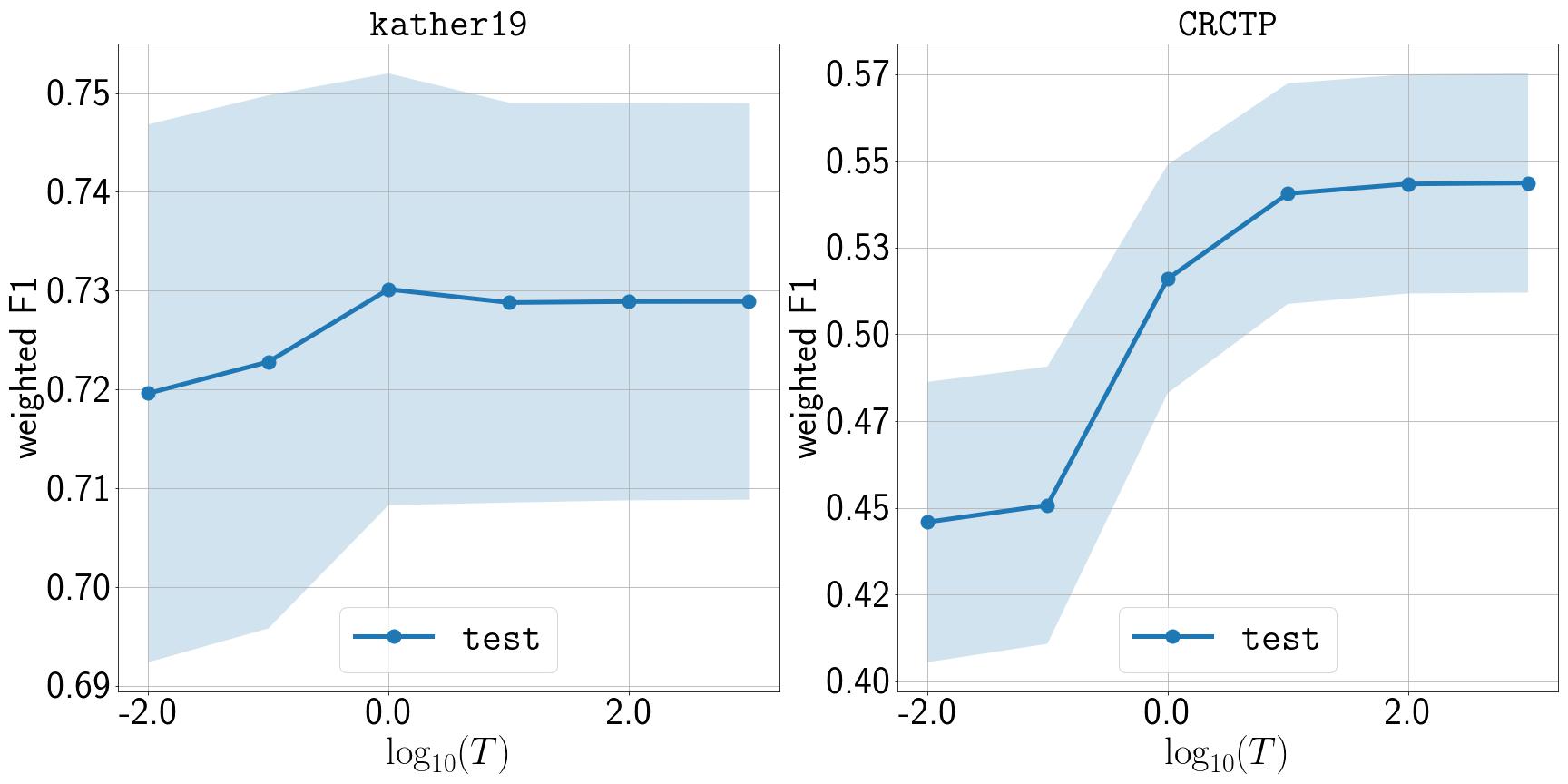}
    \caption{\update{Performances of} \textit{discriminator based top-k ensembling} with respect to $k$ (1\textsuperscript{st} row) and \textit{discriminator based weighted ensembling} with respect to $T$ (2\textsuperscript{nd} row) for Camelyon17 WILDS, Kather19 and CRC-TP datasets.}
    \label{fig:sensivity_k_T}
\end{figure}
The naive ensembling has slightly better results than the two other proposed ensembling strategies. This seems counter-intuitive but two reasons might explain this result. Firstly, domain discriminators learned during the StarGanV2 training may not be adapted for the classification tasks. Secondly, the small number of WSIs in each training dataset ($10$ and $30$) might be problematic: selecting a subset of projected images predictions when their number is already small, even if images are not well projected, might remove some of the advantages of standard ensembling. 
To explore this question, performances of discriminator based top-$k$ ensembling and weighted ensembling with respect to their hyperparameters $k$ and $T$ are reported on \autoref{fig:sensivity_k_T}. A small temperature $T$ corresponds to top-$1$ ensembling while a large $T$ is equivalent to naive ensembling or top-$k$ with the biggest possible $k$.

On the four target datasets (\texttt{val}, \texttt{test} splits of Camelyon17, Kather19 and CRC-TP), the general trend is that the more predictions we consider in the top-$k$ ensembling or the higher the temperature $T$ for the weighted ensembling the better the performances are. Increasing $k$ or $T$ in the first steps enables significant gains while for higher values the performances continue to improve or decrease only slightly. It should be noted that the decrease in performances occurs on the Camelyon17 \texttt{test} split where the number of domains is the largest in our experiments ($30$). In this case, the naive ensembling does not lead to the best performance and it seems that predictions of badly projected images decrease the performance of the ensembling method. This observation suggests that when a bigger number of available domains (WSIs) is available, the top-k and weighted ensembling methods might generalize better. 

\update{Currently, when dealing with many domains our TTA can face a computational overhead. In \update{Table 2 of supplementary material}, time complexity, scalability of the approach are investigated and suggestions are proposed to limit such overhead.}

\update{
\subsubsection{Acknowledgments}
This work was partially supported by the ANR Hagnodice ANR-21-CE45-0007. Experiments have been conducted using HPC resources from the \href{http://mesocentre.centralesupelec.fr/}{“Mésocentre”} computing center of CentraleSupélec and École Normale Supérieure Paris-Saclay supported by CNRS and Région Île-de-France. }

\section{Conclusion}
To tackle problems of domain generalization in histopathology, we have proposed a new TTA technique based on the multi-domain image-to-image translation model StarGanV2 and explored three different ensembling strategies. Experiments have been conducted on several histopathology datasets where the method has proven to be more efficient than previous color augmentation\update{/normalization}, TTA and DG methods. Even in the low data regime and when WSIs are coming from a single hospital, our method still performs well. Finally, since the method is performed at test-time, it can be combined with any other DG or data augmentation techniques applied at training-time. \update{In the future, we plan to explore StarGanV2 training on unrelated histopathology data (multi organs/cancers) and exploit it only for train-time data augmentation to check if even better protocol invariance could be achieved.}


\bibliographystyle{splncs04}
\bibliography{bibliography}

\newpage

\title{Test-time image-to-image translation ensembling improves out-of-distribution generalization in histopathology \\ (Supplementary material)}
\titlerunning{Test-time image-to-image translation ensembling}
%


\author{Marin Scalbert\inst{1, 2} \and
Maria Vakalopoulou\inst{1} \and
Florent Couzinié-Devy\inst{2}}
\authorrunning{M. Scalbert et al.}
\institute{MICS, CentraleSupélec - Université Paris-Saclay, Gif-sur-Yvette, France \\
\email{\{marin.scalbert, maria.vakalopoulou\}@centralesupelec.fr}
\and
VitaDX International, Paris, France \\
\email{f.couzinie-devy@vitadx.com}}


%
\maketitle

\begin{figure}[!htb]
    \centering
    \includegraphics[width=0.48\textwidth]{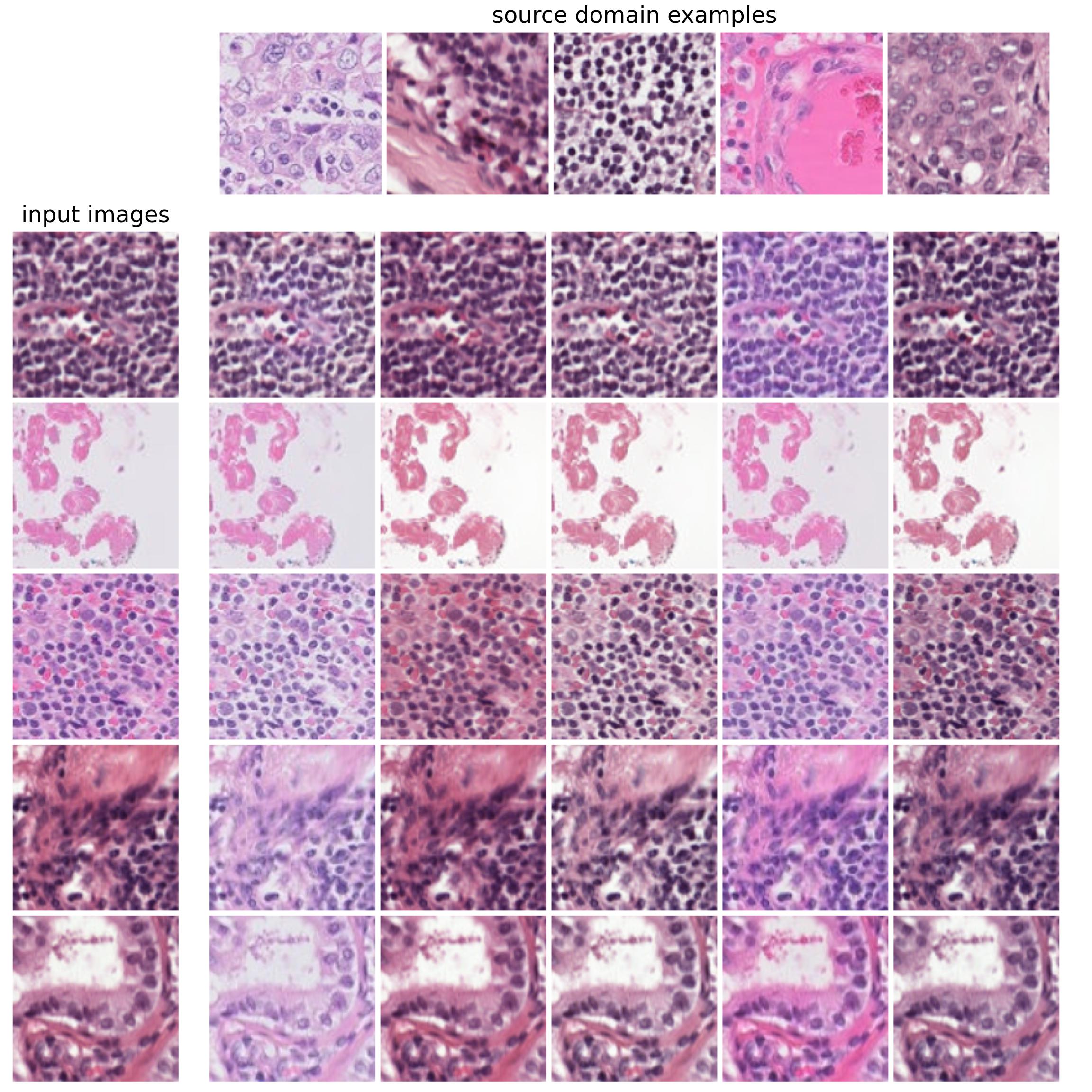}
    \includegraphics[width=0.48\textwidth]{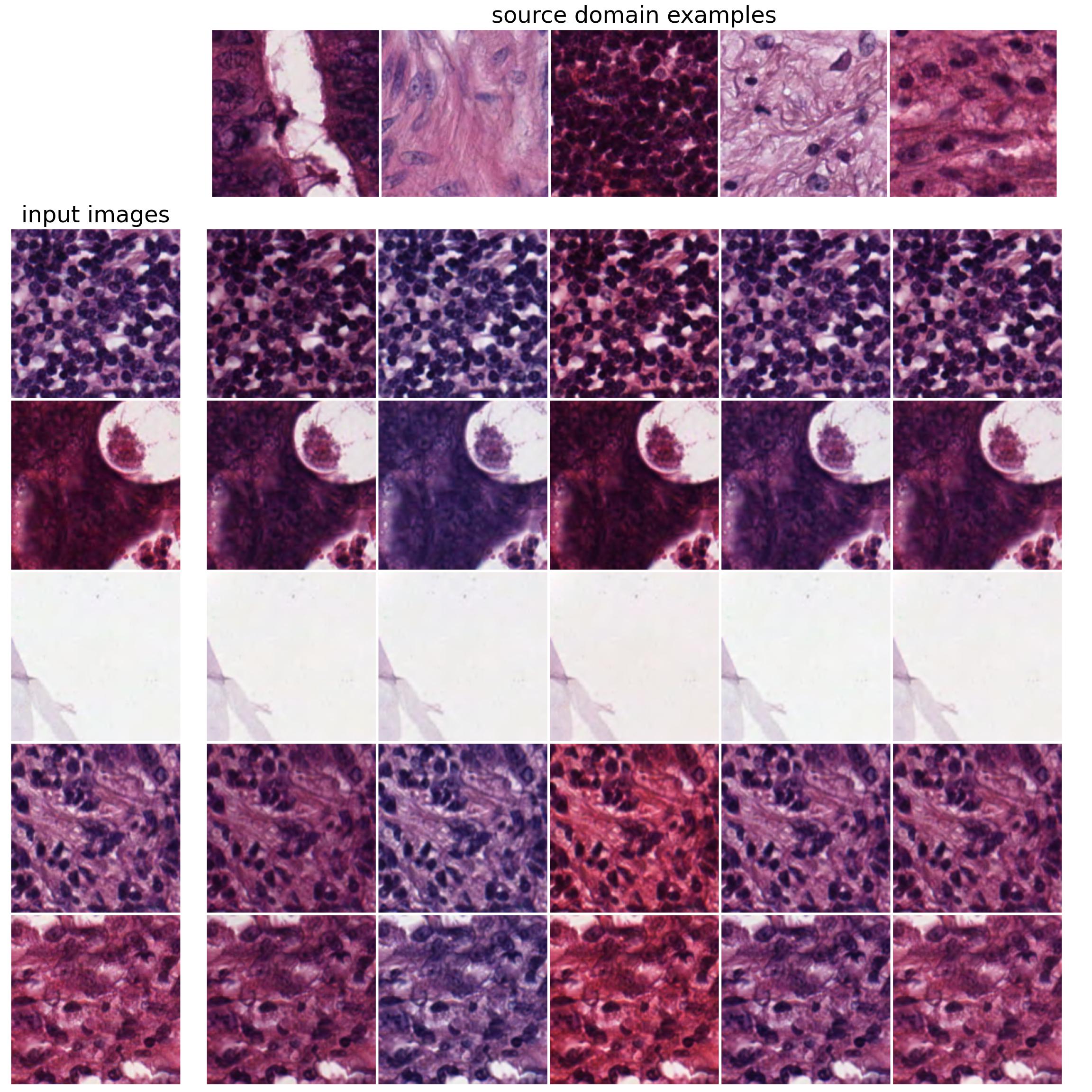}
    \caption{Translated images examples from Camelyon17 (left grid) and Kather16 (right grid). In each grid, the first column shows input images fed to StarGanV2 and the first row represents random examples from 5 target domains.}
    \label{fig:latent_guided_examples}
\end{figure}

\begin{table}[!htb]
        \centering
        \resizebox{0.8\columnwidth}{!}{
        \begin{tabular}{|l|c|c|}
        \hline
                                &  Camelyon17   &   Kather16    \\
        \hline
            image resolution    &  (128, 128)   &  (256, 256)   \\
                  $S$           &      30       &      10       \\
         style vector dimension &      64       &      64       \\
         $\lambda_{adv}, \lambda_{cyc}, \lambda_{ds}, \lambda_{percep}, \lambda_{sty}, \lambda_{gp}$     &       1       &       1       \\
               batch size       &      96       &      64       \\
             training steps     &      50K      &      40K      \\
               optimizer        &     Adam      &     Adam      \\
             learning rate      &   $1e^{-4}$   &   $1e^{-4}$   \\
                  GPUs          & 4 Nvidia V100 & 4 Nvidia V100 \\
        \hline
        \end{tabular}
        \begin{tabular}{|l|c|c|}
        \hline
                          &  Camelyon17   &   Kather16    \\
        \hline
         image resolution &  (128, 128)   &  (256, 256)   \\
              model       &  DenseNet121  &   ResNet18    \\
           pretraining    &     None      &   ImageNet    \\
            batch size    &      32       &      32       \\
          training steps  &      40K      &      4K       \\
            optimizer     &     Adam      &     Adam      \\
          learning rate   &   $1e^{-4}$   &   $1e^{-4}$   \\
               GPUs       & 1 Nvidia V100 & 1 Nvidia V100 \\
        \hline
        \end{tabular}}
        \caption{StarGanV2 (left table) and classifier (right table) hyperparameters for Camelyon17 and Kather16 datasets.}
\end{table}

\begin{figure}[!htb]
    \centering
    \includegraphics[width=0.7\textwidth]{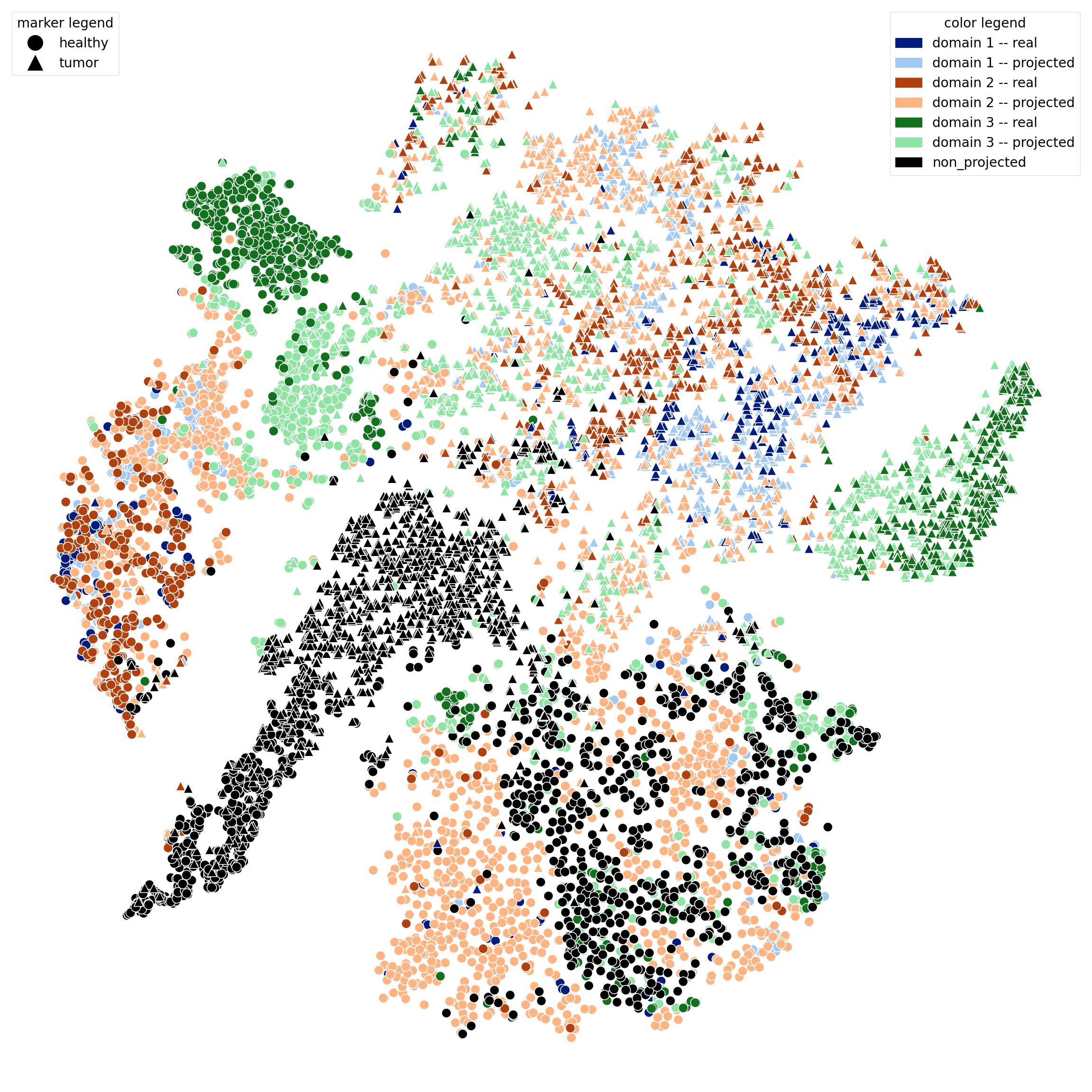}
    \caption{t-SNE of classifier last features representation for Camelyon17 \texttt{test} split and 3 random source domains. 
    Non projected target examples (black) tend to be isolated. However, when projected into a specific source domain (light color), they come closer to real examples from the domain (darker color). Additionally, this projection seems to preserve the class label: examples are still clustered by class labels (healthy/tumor).} 
    \label{fig:tsne}
\end{figure}
\begin{table}[!htb]
\centering
\resizebox{0.6\textwidth}{!}{
\begin{tabular}{|l|c|}
\hline
                    TTA method   &   Inference time per batch(ms) \\
\hline
 No TTA                &                       15.7 \\
 Geometric TTA         &                       69.7 \\
 H\&E Color Jitter TTA &                      144.0   \\
 StarGanV2 TTA         &                      667.0   \\
 Lighter StarGanV2 TTA &                      105.6 \\
\hline
\end{tabular}}
\caption{Inference times per batch for different TTA methods on Camelyon17 WILDS images with a 32 batch size and 10 domains. Geometric and H\&E Color Jitter TTA are indeed faster: our method is approximatively and respectively 10 and 4 times slower. However, our TTA leads much larger performances gains. Computational overhead of the method could be mitigated by using a more modern generator architecture without modifying the method (Lighter StarGanV2 TTA is 6 times faster) or by grouping the WSIs based on their protocol.
}
\end{table}
\end{document}